\documentclass[11pt,a4paper]{article}
\usepackage[hyperref]{emnlp2020}
\usepackage{times}
\usepackage{latexsym}
\usepackage{graphicx}
\usepackage{tabularx}
\usepackage{url}
\usepackage[inline]{enumitem}
\usepackage{array}
\usepackage{setspace}
\usepackage{multirow}
\usepackage{multicol}
\usepackage{hyperref}
\usepackage{framed}
\usepackage{color}
\usepackage{xcolor}
\usepackage{xspace}
\usepackage{amsmath}
\usepackage{amssymb}
\usepackage{cleveref}
\usepackage{booktabs}
\usepackage{colortbl}
\usepackage{lscape}
\usepackage{rotating}
\usepackage{arydshln}
\usepackage{makecell}
\usepackage{soul}
\usepackage[normalem]{ulem} 
\usepackage{subcaption}
\usepackage{soul}
\usepackage{tikz}
\usetikzlibrary{calc}
\usepackage{amsmath}
\usepackage{xcolor}
\usepackage{diagbox}
\usepackage[inline]{enumitem}

\input{special_commands.tex}
\newcommand{\eg}{\textit{e.g.}}
\newcommand{\ie}{\textit{i.e.}}

\newcommand{\bs}{\textsc{BS}}
\newcommand{\greedy}{greedy}
\newcommand{\tk}{top-k}
\newcommand{\tp}{top-p}
\newcommand{\dbs}{\textsc{DelayedBS}}
\newcommand{\gdbs}{\textsc{GroupBS}}
\newcommand{\sdbs}{\textsc{SiblingBS}}

\newcommand{\support}{\textsc{Supported}}
\newcommand{\reject}{\textsc{Refuted}}
\newcommand{\nei}{\textsc{Not Enough Info}}

\newcommand{\ccnews}{\textsc{CCNews}}
\newcommand{\openwebtext}{\textsc{OpenWebText}}
\newcommand{\bookcorpus}{\textsc{Books}}
\newcommand{\wikipedia}{\textsc{Wikipedia}}

\newcommand{\fairseq}{\textsc{fairseq}}

\usepackage{microtype}

\aclfinalcopy 

\newcommand{\sapienza}{$^1$}
\newcommand{\fair}{$^2$}
\newcommand{\ucl}{$^3$}
\newcommand{\uclfair}{$^{2,3}$}

\title{How Decoding Strategies Affect the Verifiability of Generated Text}

\author{
Luca Massarelli\sapienza\thanks{~~Work done during internship with Facebook.} \thanks{~~Equal contribution.} \ ~Fabio Petroni\fair{}\samethanks \ \quad Aleksandra Piktus\fair{}\samethanks \ \quad Myle Ott\fair{} \\
{\bf   Tim Rockt\"aschel\uclfair{} \ Vassilis Plachouras\fair{} \ Fabrizio Silvestri\fair{} \ Sebastian Riedel\uclfair{} }\\
\sapienza{} Sapienza University of Rome \\
{\tt massarelli@diag.uniroma1.it}\\
\fair{}Facebook AI \\
{\tt \{fabiopetroni, piktus, myleott, rockt,} \\ 
{\tt \quad\quad\quad\quad vplachouras, fsilvestri, sriedel\}@fb.com} \\
\ucl{}University College London\\
}

\date{}

\begin{document}
\maketitle
\begin{abstract}
  Recent progress in pre-trained language models led to systems that are able to generate text of an increasingly high quality. 
While several works have investigated the fluency and grammatical correctness of such models, it is still unclear to which extent the generated text is consistent with factual world knowledge.
Here, we go beyond fluency and also investigate the verifiability of text generated by state-of-the-art pre-trained language models.
A generated sentence is verifiable if it can be corroborated or disproved by Wikipedia, and we find that
the verifiability of generated text strongly depends on the decoding strategy. 
In particular, we discover a tradeoff between factuality (i.e., the ability of generating Wikipedia corroborated text) and repetitiveness.
While decoding strategies such as top-k and nucleus sampling lead to less repetitive generations, they also produce less verifiable text. 
Based on these finding, we introduce a simple and effective decoding strategy which, in comparison to previously used decoding strategies, produces less repetitive and more verifiable text.
\end{abstract}

\section{Introduction}

Recent years have led to a considerable surge of interest in and capabilities of pre-trained language models (LMs). Today, they play a critical role in many NLP tasks, such as text classification, machine comprehension and natural language inference \cite{DBLP:conf/naacl/PetersNIGCLZ18,  DBLP:journals/corr/abs-1810-04805,liu2019roberta, yang2019xlnet}, to name just a few. They serve as a pre-training objective for downstream applications and they have been used to showcase and measure the general progress in NLP \cite{DBLP:conf/iclr/YuBDGK17, liu2019rescoring}.

  Several works \cite{radford2019language,DBLP:journals/corr/abs-1909-05858} show the remarkable fluency and grammatical correctness of text decoded from modern LMs. Additionally, recent works \cite{petroni2019language,logan2019barack, Broscheit2019Investigating, roberts2020much} demonstrate that beyond general linguistic capabilities, language models can also pick up factual knowledge present in the training data. However, it is unclear if LMs are able to convey such knowledge at decoding time when producing long sequences---do they generate fluent, grammatical but 
 ``babbler-level'' text or can they produce utterances that reflect factual world knowledge? 
  
Understanding this behaviour becomes crucially important as the downstream adoption of automatically generated text increases. 
Already today LMs face growing scrutiny from the media and the broader society, as well as from the researchers themselves.
For example, \citet{radford2019language} initially argued against releasing their models in order to prevent automatic generation of fake news~\cite{openai2019better}. Several blogs and web resources demonstrate that differentiating between human and machine-generated text has become surprisingly difficult.\footnote{\url{http://quiz.newsyoucantuse.com/}}

\begin{figure*}[t]
    \centering
    \includegraphics[width=\linewidth]{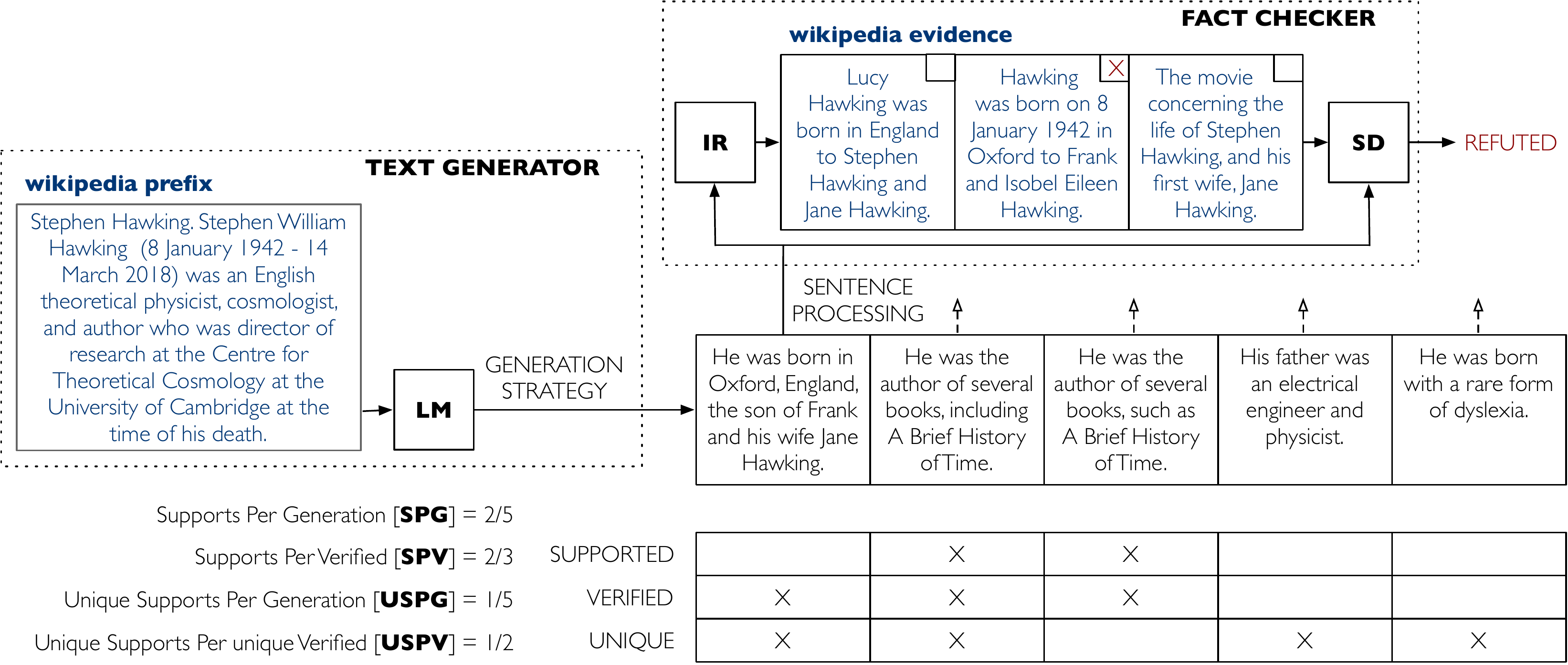}
    \caption{High level description of our experimental methodology that combines a language model (LM) with a fact checker, usually implemented combining an information retrieval (IR) and a stance detector (SD) component. 
    }
    \label{fig:high_level_desc}
\end{figure*}

With that in mind, we set out to study state-of-the-art auto-regressive transformer-based language models through the lens of their verifiability. Specifically, we use Wikipedia to first create a set of natural language prompts to initiate generation. Next, we use transformer models of various sizes and trained with different corpora to generate sentences off these prompts with varying decoding configurations. 
Finally, following earlier work in fact checking~\cite{Thorne_2018}, we use Wikipedia again to verify each sentence as supported, refuted, or unverifiable using both an off-the-shelf automatic fact-checking system and human annotators. 
We define verifiability metrics on top of the automatic and human fact-checkers' evaluation outcomes (see Figure \ref{fig:high_level_desc} for a high-level overview). 

The truthfulness of generated text can be traded off with other properties. 
For example, a decoding algorithm can generate the same true fact over and over again to produce many verifiable utterances, but this would be a poor outcome in terms of repetitiveness. 
Similarly, a model might generate ungrammatical text that cannot be verified as supported or refuted at all, and hence not as factually wrong either. 
Our experiments show that the text generated from auto-regressive transformer-based LMs, especially in their large versions (1.4B parameters), is almost always grammatical and fluent regardless of the configuration, but that repetitiveness can vary a lot. We hence focus on this dimension in our analysis and define metrics that combine repetitiveness with verifiability. 

One of our main findings is that while sampling methods, such as top-k and nucleus, produce more natural and less repetitive text, they also generate fewer supported and more refuted statements.
Beam search, on the other hand, shows much better performance along these dimensions at the cost of producing highly repetitive text. Based on these observations, and inspired by findings in~\citet{holtzman2019curious}, who showed how the probability of human text under language models is varying from token to token, we introduce a simple strategy: Delayed Beam Search (\dbs). 
In \dbs{}, we iterate between sampling and finding most likely utterances.
By simply injecting stochasticity in the beginning of a sentence and then switching to beam search, we generate text that is less repetitive while at the same time scores well in terms of our verifiability metrics. Our main findings hold across several experimental settings, with varying training set size and model size.

To summarize, we make the following contributions:
\begin{enumerate*}[label=(\roman*)] 
\item we propose an experimental methodology to assess machine generated text with respect to repetitiveness and verifiability.
\item we assess a wide range of decoding algorithms with respect to these dimensions,
\item we introduce a novel decoding strategy that addresses some of the shortcomings of existing solutions,
\item we carry out an annotation campaign to validate our findings and assess the quality of the automatic fact checking system.
\end{enumerate*}
\section{Related Work}
\citet{DBLP:journals/corr/abs-1909-05858} trained CTRL, a large (1.63B parameters) pretrained language model that can be conditioned on style or content for controlling generated text. Users can, for example, specify the domain, entities, as well as relationships between entities, to control the generated text. While impressive, their work does not provide insights into the verifiability of the generated text.

Multiple efforts focus on improving text decoding with respect to different criteria.
\newcite{vijayakumar2016diverse} and \newcite{li2016simple} introduce alternative scoring strategies to diversify the hypothesis tree explored by beam search. \citet{DBLP:conf/acl/LewisDF18} propose \emph{top-k sampling}, \ie{}, sampling from the top k tokens with the highest probability to generate stories. \citet{holtzman2019curious} find that for the same neural language model, the choice of the decoding strategy can have a dramatic effect on the fluency and repetitiveness of the generation. They propose \emph{nucleus sampling} as a way to increase diversity of the generated text while improving fluency. In our work, we find that while this strategy does create more fluent and less repetitive text, it does also result in a less factually true generation. \newcite{cho2019diverse} choose to separate the generation and diversification steps altogether, and focus on leveraging content selection to map the input to diverse sequences. We describe various generation strategies in more detail in section \ref{sec:Background}.

\citet{DBLP:journals/corr/abs-1908-04319} note that with nucleus sampling, per-token probabilities can be very low which they attribute to the likelihood training objective. They propose a novel \emph{unlikelihood} training objective which lowers the probability of tokens in the context of the model. Their approach is orthogonal to the decoding strategy and testing alternative training objectives is out of the scope of our paper.

A recent approach by \newcite{bakhtin2019real} learns to distinguish human from machine generated text. \citet{DBLP:journals/corr/abs-1905-12616} investigate generating and detecting fake news using neural language models. \citet{niewinski2019tmlab} propose a variation of the GPT-2 language model to explicitly generate malicious claims.
Instead of directly optimizing for generating fake or factual news, we are interested in investigating the relationship between the verifiability of the existing language models and different decoding strategies they are coupled with.

Several metrics have been proposed to evaluate natural language generations in the past \cite{novikova2017we}. Given that recent studies \cite{DBLP:conf/acl/LewisDF18,holtzman2019curious,DBLP:journals/corr/abs-1908-04319} point to repetitiveness as one of the main problems affecting the generation of state-of-the-art models, we mainly consider this dimension in our analysis. 
\section{Background}
\label{sec:Background}

Language models (LMs) assign probabilities to sequences of tokens. Given a context, that is, a sequence of tokens $\mathbf{c_t} = [w_1, w_2, \ldots, w_{t-1}]$, autoregressive LMs commonly estimate the probability distribution of the next target using neural models~\citep{DBLP:conf/slt/MikolovZ12,DBLP:journals/corr/MelisDB17,DBLP:journals/jmlr/BengioDVJ03} with:
\begin{equation}
    p(w_t\,|\,c_{t}) = \softmax(\mathbf{W}\mathbf{h}_t+\mathbf{b})
    \label{eq:toplayer}
\end{equation}
where $\mathbf{h}_t \in \mathbb{R}^k$ is the output vector of a neural network at position $t$ and $\mathbf{W} \in \mathbb{R}^{|\mathcal{V}|\ \times\ k}$ is a learned parameter matrix that maps $\mathbf{h}_t$ to unnormalized scores for every word in the vocabulary $\mathcal{V}$.
In this work, we consider self-attention mechanisms \citep{radford2018improving,DBLP:journals/corr/abs-1901-02860,radford2019language} to compute $\mathbf{h}_t$ given the word history.

\paragraph{Open-Ended Text Generation}
As described in \citet{holtzman2019curious}, the task of open-ended text generation involves producing a coherent completion of the provided context. We consider the common left-to-right generation, where a token at position $t$ in the sequence is generated by considering the probability distribution over the vocabulary defined in equation \ref{eq:toplayer}. 
Once a decision is made for $w_t$ according to a decoding strategy, it is incorporated into the context and the process is iterated - \ie, the token at position $t+1$ is generated by considering $p(w_{t+1}\,|\,c_{t+1}=[w_1, \ldots, w_{t}]$). In this work, we consider different decoding strategies of selecting $w_t$ given $p(w_t\,|\,c_{t})$.

\subsection{Decoding Strategies}
\label{generation_strategies}

The decoding strategies we consider in our analysis can be broadly divided in two families: \textit{sampling-based} and \textit{likelihood-based}. 

\paragraph{Sampling-based}
This family of techniques aims at increasing the diversity of the output and avoiding repetitions by introducing stochastic decisions during the generation process.

\noindent
\textit{Top-k sampling}~\cite{DBLP:conf/acl/LewisDF18} selects $w_t$ by sampling from the $k$ tokens with the highest probability in $p(w_t\,|\,c_{t})$.

\noindent
\textit{Top-p sampling}, also referred to as nucleus sampling~\cite{holtzman2019curious}, selects $w_t$ from the smallest set of tokens whose cumulative probability (given by $p(w_t\,|\,c_{t})$) is above a threshold $p$.

\paragraph{Likelihood-based}
These strategies navigate the solution space by selecting sequences of tokens that maximize the overall likelihood.  Given that the number of possible sequences is typically very large, it is a common practice to define heuristics to make the generation practical.

\noindent
\textit{Beam Search} (\bs). This strategy approximately maximizes the likelihood of the whole sequence. Throughout the generation, we hold a \textit{beam} of $\beta$ prefixes which are iteratively extended. At each time-step, $\beta$ tokens are generated to complete each of the prefixes in the beam and we retain $\beta$ hypotheses with the highest score out of the $\beta^2$ candidates for the next step. $\beta$ is referred to as the beam size. \textit{Greedy} decoding, where at each step the most likely token is selected, is a special case of beam search with beam size 1.

\noindent
\textit{Group diverse Beam Search} (\gdbs). To favor the diversity of the exploration, \citet{vijayakumar2016diverse} propose to divide the beam into groups. The diversity between groups is imposed by introducing a group dissimilarity penalty into the search objective.

\noindent
\textit{Sibling diverse Beam Search} (\sdbs). With the same aim of diversifying the exploration, \citet{li2016simple} propose a variant of beam search which introduces a penalty proportional to the rank of a candidate token with respect to its source in the beam. The goal is to encourage preserving hypotheses from diverse sources within the beam.

A simple trick to reduce repetitiveness is to explicitly prevent the generation of already observed $n$-grams \cite{paulus2017summarization}. We refer to this approach as \textit{n-gram blocking}.

\paragraph{Delayed Beam Search (\dbs).}
We propose a new hybrid strategy that uses sampling to generate the first $L$ tokens of a sentence and then it finishes the sentence using beam search. The smaller the $L$, the closer the behaviour is to beam search. Conversely, the larger the $L$, the closer we are to sampling strategies. Consequently, by tuning $L$, it is possible to combine the advantages of both sampling and likelihood-based strategies.

\section{Evaluating Verifiability}
\label{sec:metrics}
In this section we first describe the tools used to evaluate the verifiability of the generated text. We then formally introduce our repetitiveness and verifiability metrics.

The high level overview of our evaluation setup is shown in Figure~\ref{fig:high_level_desc}. For the purpose of this analysis, we consider both the \textit{text generator} and the \textit{fact checker} as black boxes which produce and assess text respectively.
More specifically, the text generator gets in input a prefix $p$ and produces a sequence of tokens that can be interpreted as a completion of $p$. We segment the generated completion into sentences and consider the first $k$ sentences. The fact checker gets in input a sentence and outputs a positive (\support), negative (\reject) or unverifiable (\nei) response as well as textual evidence used for the judgment. We consider a sentence as \textit{verified} if the output label is either \support{} or \reject.

Our metrics assess the generation process given a set of prefixes $P$. The set $P$ can be seen as the data source for our verifiability probe.
Let $G^p = [s^p_1, ..., s^p_k]$ be the sequence of sentences generated by the LM from prefix $p \in P$. 
We indicate with $V^p \in G^p $ the set of sentences that are verified by the fact checker, while with $S^p \in V^p$ we denote the subset of sentences labeled as \support{}.
To assess the verifiability of the generated text we introduce the following two metrics:

\noindent
\textbf{Supports Per Generation (SPG)}: is the fraction of supported sentences among the generated ones:
\begin{equation}
    \text{SPG} =\frac{1}{|P|}\sum_{p \in P}{\frac{|S^p|}{k}}
\end{equation}

\noindent
\textbf{Supports Per Verified (SPV)}: is the fraction of supported sentences among the verified ones:
\begin{equation}
    \text{SPV} =\frac{1}{|P|}\sum_{p \in P}{\frac{|S^p|}{|V^p|}}
\end{equation}
\noindent
SPG can be interpreted as a sort of a recall metric while SPV as a precision one.

Note that a generation could achieve a high score in terms of SPG and SPV by repeating the same supported sentence over and over again. To capture this behaviour, we define the \textit{unique} variants of our metrics. We consider two sentences as equivalent if they have the same factuality label (\ie, \support{} or \reject{}) and the decision is justified by the same evidence. For a set of equivalent sentences, we consider only the one which appeared first in the generation as unique. We denote the set of unique sentences as $S_u^p \in S^p$, $V_u^p \in V^p$ is a set of unique verified sentences. We introduce:

\noindent
\textbf{Unique Supports Per Generation (USPG)}: the fraction of unique supported sentences among the generated ones:
\begin{equation}
    \text{USPG} =\frac{1}{|P|}\sum_{p \in P}{\frac{|S_u^p|}{k}}
\end{equation}{}

\noindent
\textbf{Unique Supports Per unique Verified (USPV)}: the fraction of unique supported sentences among unique verified sentences:
\begin{equation}
    \text{USPV} =\frac{1}{|P|}\sum_{p \in P}{\frac{|S_u^p|}{|V_u^p|}}
\end{equation}{}
\section{Methodology}
\label{sec:method}

In this section we describe in detail the implementational choices for all components in Figure~\ref{fig:high_level_desc}.

\paragraph{Prefix Dataset}

We retrieve title and description of the top-1000 most visited Wikipedia pages of 2017 and 2018.
For each page, we concatenate the title and the first sentence in the description to create a string prefix for the language model.
We use 2018 data as validation set and run parameter sweeps over it. We tested the best configuration of every decoding strategy on 2017 data (test set). We ensure no overlap between 2017 and 2018 prefixes. 

\paragraph{Language Model}

We consider three sizes of language models (small, medium, large) based on the Transformer architecture~\cite{DBLP:conf/nips/VaswaniSPUJGKP17,radford2019language}, with 124M, 354M and 1.4B parameters respectively.
We train models on four corpora: 
\begin{enumerate*}[label=(\roman*)]
\item \wikipedia{}, an English Wikipedia dump consisting of roughly 2 Billion Words;
\item \bookcorpus{}, the Toronto books corpus~\cite{zhu2015aligning,kiros2015skip}, which consists of fiction books totaling about half a billion words;
\item \openwebtext{}, a reconstruction of the WebText corpus~\cite{radford2019language} consisting of roughly 3 Billion Words; \
\item \ccnews{}, a de-duplicated subset of the English portion of the CommonCrawl news
dataset \cite{Nagel2016,bakhtin2019real,liu2019roberta}, which totals around 16 Billion words.
\end{enumerate*}
We train models using the \fairseq{} toolkit~\cite{ott2019fairseq}.

\paragraph{Generation Strategy}

We consider the generation strategies discussed in Section~\ref{generation_strategies}, namely \tk{}, \tp{}, \greedy{}, Beam Search (\bs{}), Group-Diverse Beam Search (\gdbs{}), Sibling-Diverse Beam Search (\sdbs{}) and Delayed Beam Search (\dbs{}). Additionally, we experiment with $n$-gram blocking and indicate that a model is equipped with blocking with a subscript $b$, \eg{}, \bs$_b$. We fix the generation length to 256 tokens. We perform three generations per prefix with different seeds for all strategies that make stochastic decisions, and report average values.

\paragraph{Sentence Processing}
Given that our fact checker expects a single sentence as input, we segment the generated text into sentences.
We consider the first $k=5$ sentences.
We perform coreference resolution to replace pronouns with the corresponding referring entity in order to give the complete information to the fact checker.
For the same reason, we apply a simple heuristic that replaces each determiner (\ie, \textit{"The"}) at the beginning of a sentence and the subsequent noun with the original entity (\ie, the title of the Wikipedia page).
For all these steps we use spaCy.\footnote{\url{https://spacy.io}} 
We consider sentences longer than 50 tokens as not verifiable, since long sentences are likely to contain multiple claims and can be misclassified by the automatic fact-checking system, we consider that has been trained on short single claim statements.

\begin{table*}[]
\centering

\resizebox{\textwidth}{!}{   

    \begin{tabular}{ c c c c c c c c c c c } 
    \multicolumn{2}{c}{\multirow{2}{*}{\diagbox[width=\dimexpr 0.15\textwidth+2\tabcolsep\relax, height=1.3cm]{strategies}{\raisebox{-1ex}{metrics}}
    }} & \phantom{x} & \multicolumn{2}{c}{\textit{repetitiveness}} & \phantom{x} & \multicolumn{2}{c}{\textit{verifiability}} & \phantom{x} & \multicolumn{2}{c}{\textit{diverse verifiability}} \\
    
    \cmidrule{4-5}
    \cmidrule{7-8} 
    \cmidrule{10-11} 
    
    &&&  \makecell{ \textbf{distinct} \\ \textbf{4-grams}}  &  \makecell{ \textbf{4-grams} \\ \textbf{proportion}} && \textbf{SPG} & \textbf{SPV} && \textbf{USPG} &  \textbf{USPV} \\
    \midrule
    
    \multicolumn{2}{c}{\textit{human - Wikipedia}}  && 222.48 & 100.00 &&  36.56 & 93.03 && 36.56	& 93.03 \\
     
    \midrule
    
    \multirow{2}{*}{\textit{sampling}} & \textbf{\tk}   && \textbf{143.52} &     \textbf{64.51} && 13.02  & 70.15  &&  11.06  & 69.39 \\
     & \textbf{\tp}   &&       136.66 &     61.43 && 13.94  & 70.76  && 11.36  & 68.93 \\
     
    \midrule
    
    \multirow{4}{*}{\textit{likelihood}}  & \textbf{\greedy}   &&             67.42 &     30.31 && 19.62  & 78.67  && 12.06  & 77.21  \\
    & \textbf{\bs}   &&            59.53 &     26.76 && \textbf{25.50}  & \textbf{84.49} && 11.88  & \textbf{81.59}  \\
    & \textbf{\gdbs}   &&             66.06 &     29.69 && 20.56  & 78.29  &&  11.54  & 76.53     \\
    & \textbf{\sdbs}   &&         67.11 &     30.16 && 22.32  & 80.11  && 11.36  & 76.76 \\
    
    \midrule
    
    \textit{hybrid} & \textbf{\dbs}   &&       112.12 &     50.40 &&   17.52  & 78.99  && 12.74  & 77.59 \\
    
    \midrule
    
     \textit{blocking}  & \textbf{\bs$_b$}   &&       92.00 &     41.35 && 23.62  & 83.35  &&  \textbf{15.28}  & 80.76  \\
    
    \bottomrule
\end{tabular}  }  
\caption{
Performance of the different generation strategies on the considered metrics. We report percentage values for the large transformer model on the test set. 
The first row shows human performance computed on Wikipedia. 
}
\label{tab:strategies}

\end{table*}

\paragraph{Fact Checker}
We consider an off-the-shelf fact checker\footnote{\url{https://github.com/dominiksinsaarland/domlin_fever}} trained on the FEVER dataset~\cite{Thorne_2018} which achieves the highest FEVER score of 68.46\% in the second FEVER shared task \cite{thorne-etal-2019-fever2}.
This solution takes inspiration from \citet{AtheneFever} and consists of three main stages: 
\begin{enumerate*}[label=(\roman*)]
\item identify relevant Wikipedia pages, as in \citet{AtheneFever};
\item retrieve relevant sentences from such pages;
\item recognize textual entailment between input and retrieved text.
\end{enumerate*}
The system uses a hierarchical sentence retrieval approach in order to verify claims that require multiple statements as evidence. It uses BERT \cite{DBLP:journals/corr/abs-1810-04805} for both retrieval and entailment.

\paragraph{Metrics}

We use all the metrics introduced in Section \ref{sec:metrics}. We also consider the following metrics to capture the repetitiveness of the generation:

\noindent
\textit{Distinct 4-gram}: is the average number of distinct 4-grams present in the generated text \cite{vijayakumar2016diverse}.

\noindent
\textit{4-gram proportion}: is the average ratio between distinct 4-grams in machine and human generated text \cite{holtzman2019curious}. For the latter, we consider the 256 tokens after the first sentence in the description for each Wikipedia page.

\section{Results}
\label{sec:results}

\begin{figure*}[!t]
    \centering
    \begin{subfigure}[b]{0.47\textwidth}
        \includegraphics[width=\textwidth]{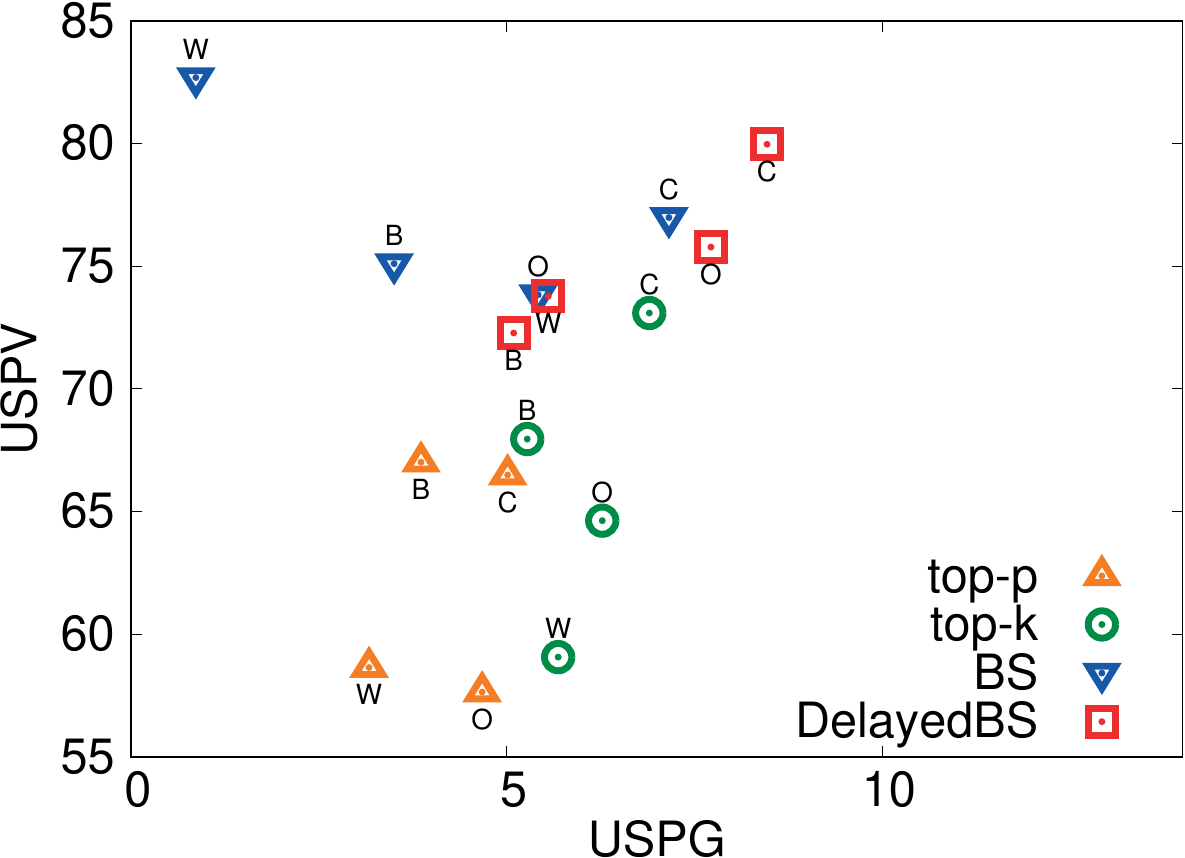}
        \caption{Performance of the small transformer model trained on different corpora, i.e., \wikipedia~(W), \bookcorpus~(B), \openwebtext~(O) and \ccnews~(C).}
        \label{fig:abldatazize}
    \end{subfigure}
    \hfill
    \begin{subfigure}[b]{0.47\textwidth}
        \includegraphics[width=\textwidth]{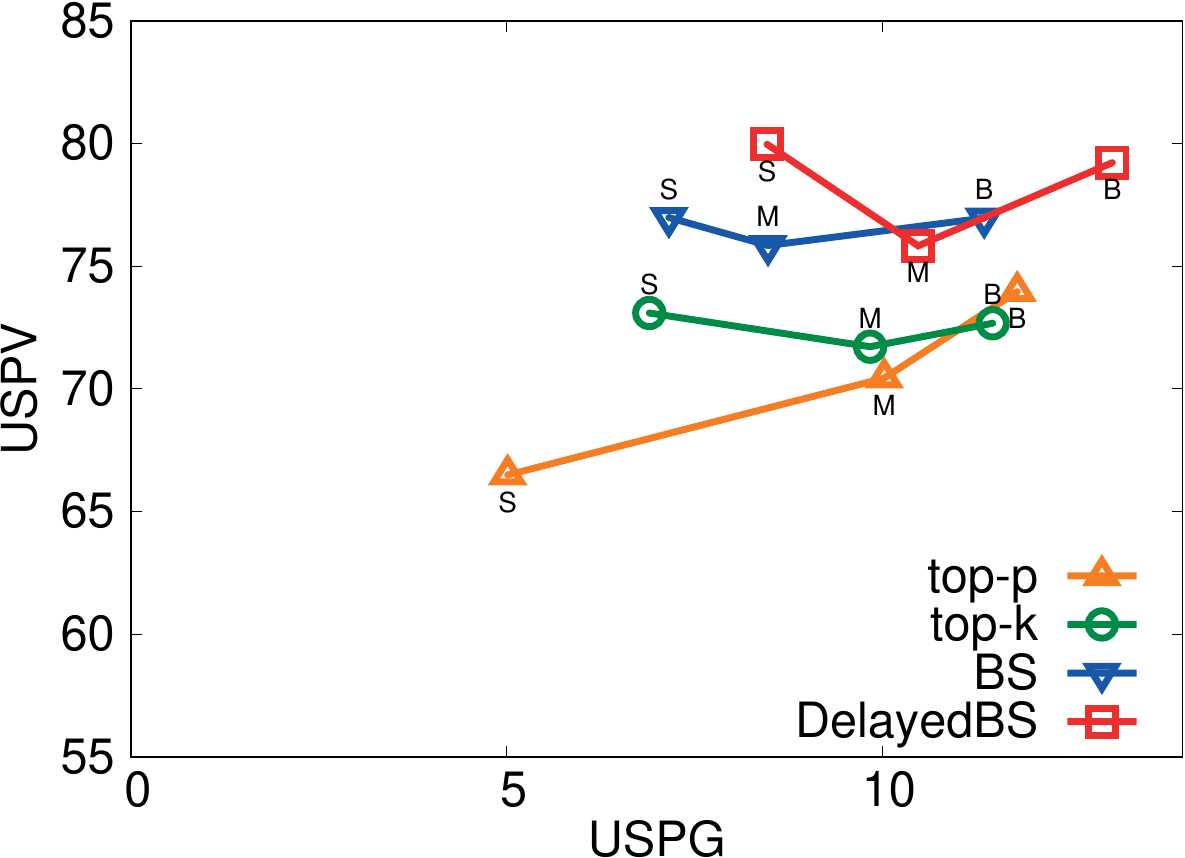}
        \caption{Ablation study on our transformer model trained on \ccnews~ with increasing number of parameters, i.e., 124M (S), 354M (M) and 1.4B (B).}
        \label{fig:ablmodzize}
    \end{subfigure}
    \caption{USPV vs USPG, inspired by precision-recall curve.}
    \label{fig:pr}
\end{figure*}

We summarize the main results in Table \ref{tab:strategies}. It shows the performance of the different generation strategies on the considered metrics on the test set of prefixes, considering the large transformer model trained on \ccnews{} (this corpus led to the best performance according to our ablation, see Figure \ref{fig:abldatazize}). 
We performed an exhaustive grid search over the parameters for all considered generation strategies using the small model on the validation set,
and consider the configuration that led to the highest USPG value (see the Appendix for details). 
We report as reference human performance computed on Wikipedia considering at most the first 5 sentences of the prefix article.

Sampling strategies (\ie, \tp{} and \tk) outperform the other strategies in terms of repetitiveness metrics, that is, they are able to generate text with a higher degree of diversity, consistently with previous works \cite{DBLP:conf/acl/LewisDF18,holtzman2019curious}. However, diversity comes at a price, as the verifiability metrics are low (in particular, precision values - they generate more refuted sentences).
Intuitively, random choices might hamper verifiability when sampling a token in specific positions of the sentence, for instance, in a named entity, potentially making the overall sentence non factual. We notice that this problem gets even worse by increasing $k$ or $p$.
Following a generation path that maximizes likelihood is a better approach for verifiability. In particular, \bs{} achieves the highest performance in terms of SPG and SPV. Nevertheless, generation diversity drops, consistently with previous works \cite{vijayakumar2016diverse, li2016simple, DBLP:journals/corr/abs-1908-04319, holtzman2019curious}. Solutions such as \gdbs{} and \sdbs{} have been proposed to mitigate this problem, and their numbers actually look slightly better than \bs{} in terms of repetitiveness metrics. 

When we assess diverse verifiability (that is, we consider distinct supported/refuted sentences), likelihood and sampling based strategies are similar in terms of recall (\ie, USPG), while likelihood-based solutions outperform both \tk{} and \tp{} in terms of precision (\ie, USPV) by a large margin - they generate less sentences refuted by the fact checker.

\begin{figure*}[!t]
    \centering
    \begin{subfigure}[b]{0.32\textwidth}
        \includegraphics[width=\textwidth]{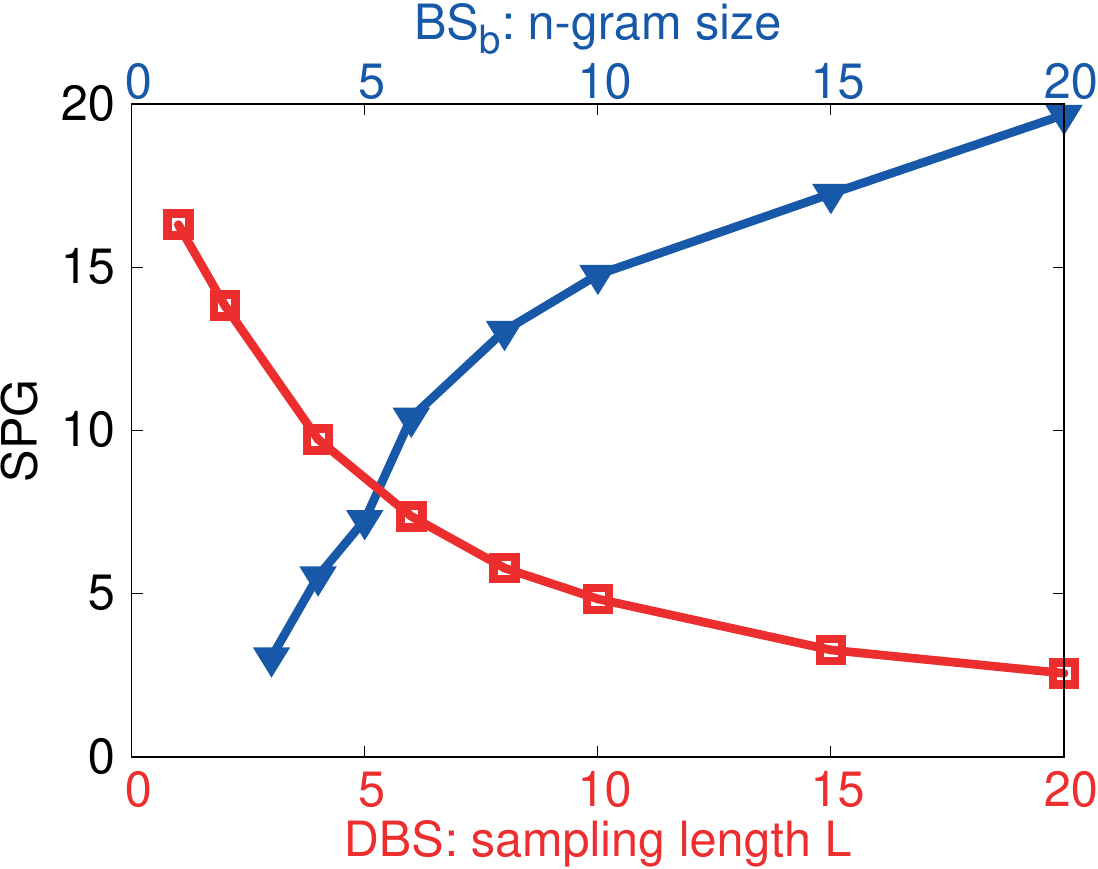}
        \label{fig:SPG}
    \end{subfigure}
    \begin{subfigure}[b]{0.32\textwidth}
        \includegraphics[width=\textwidth]{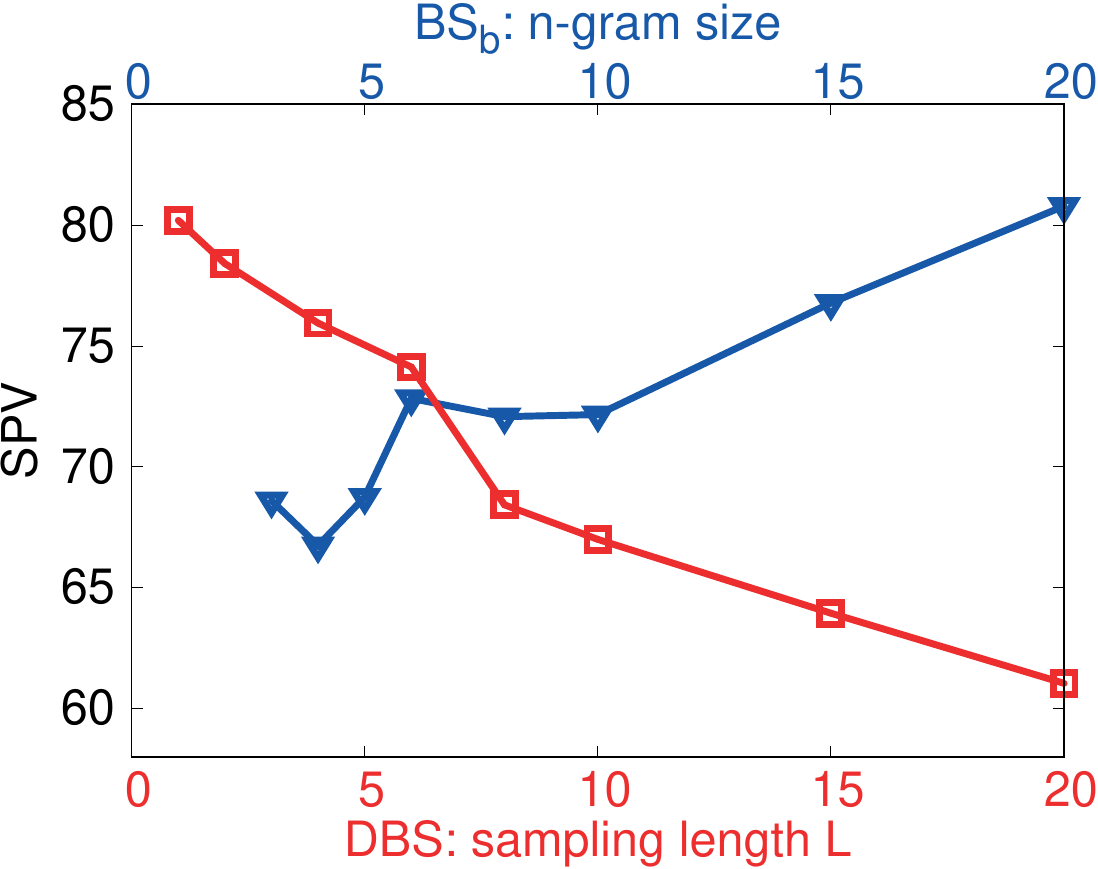}
        \label{fig:SPV}
    \end{subfigure}
    \begin{subfigure}[b]{0.32\textwidth}
        \includegraphics[width=\textwidth]{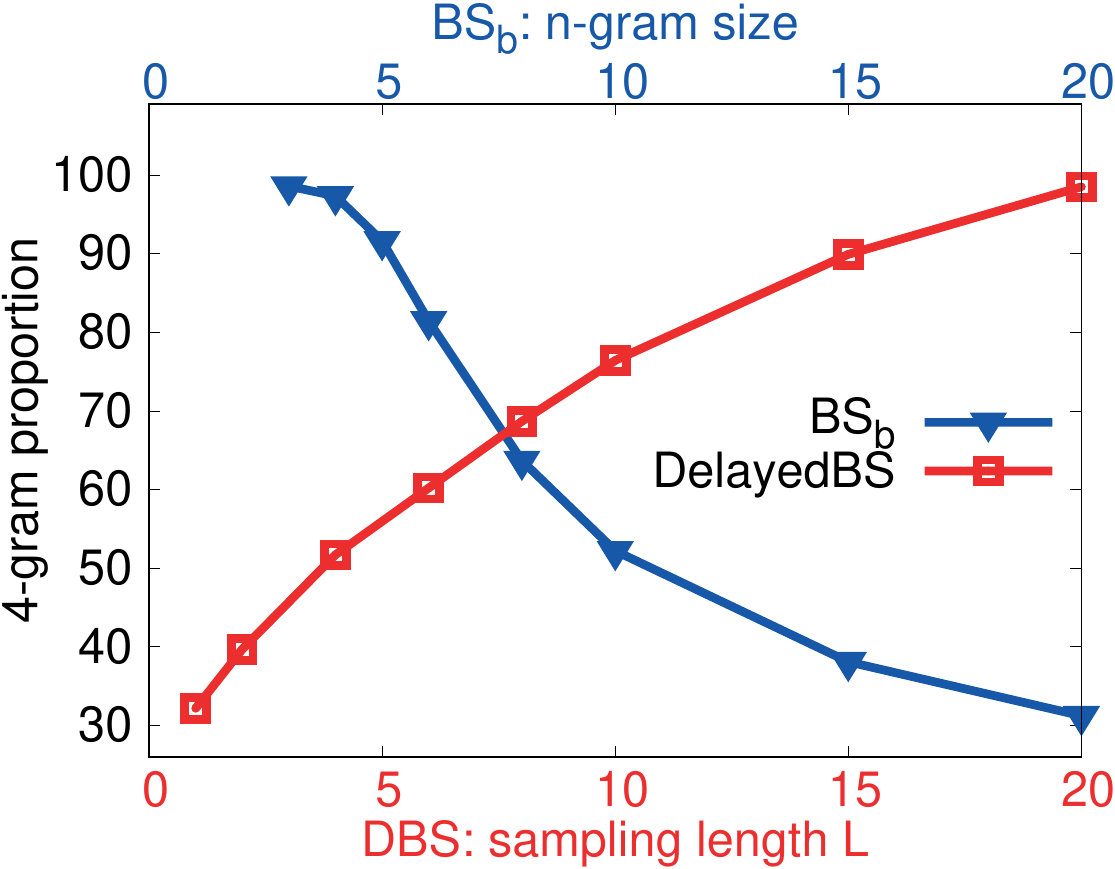}
        \label{fig:4gram}
    \end{subfigure}
    \caption{SPG, SPV and 4-gram proportion values for \bs$_b$ and \dbs{}, by varying the sampling length $L$ for \dbs{} (bottom axis) and the n-gram blocking size for \bs$_b$ (top axis).}
    \label{fig:tradeoff}
\end{figure*}

\dbs{} tries to combine the best of these two approaches, by defining a hybrid strategy that starts a sentence by sampling tokens and ends it by following a max-likelihood path. It achieves results comparable to likelihood-based solutions in terms of precision and recall for diverse verifiability while being much less repetitive (it almost doubles the number of distinct 4-grams). Interestingly, it is sufficient to sample just the first token with high uncertainty (top-100) and finish the sentence with beam search to trigger this behaviour (Figure \ref{fig:ablationdelaylenght} in the Appendix Section reports a detailed ablation study for the delay length).

Another way of mitigating repetitiveness is through $n$-gram blocking. We combine it with \bs{}, sweeping over the values of $n$ between 3 and 20. In line with our expectations, low $n$ values score low in verifiability metrics, as the model is forced to explore less likely parts of the solution space in order to avoid generating previously observed $n$-grams. Unsurprisingly, the diversity of the solution drops as $n$ increases. In this sense, \bs$_b$ and \dbs{} attempt to strike a similar balance between diversity (introduced via $n$-gram blocking in \bs$_b$ and via sampling in \dbs{}) and verifiability (achieved by incorporating \bs{}). Figure~\ref{fig:tradeoff} highlights this analogy further. Overall, we achieve the best USPG performance by combining 20-gram blocking and \bs{} - we believe it is due to the fact that $n$-gram blocking prevents \bs{} from repeating the same phrases multiple times, while remaining relaxed enough to allow the generation to produce a high-likelihood solution. However, even though \bs$_b$ archives the best results in terms of diverse verifiability metrics, \dbs{} still produces less repetitions, hence constituting a viable alternative.

\paragraph{Ablation studies} We experiment with different training corpora (Figure \ref{fig:abldatazize}) and different sizes of the transformer model (Figure \ref{fig:ablmodzize}), using the validation set. We report USPV vs USPG values, taking inspiration from the popular precision-recall curve. The average perplexity of the small transformer model is the lowest for \wikipedia{} (8.31) compared to \bookcorpus{} (53.08), \openwebtext{} (11.14) and \ccnews{} (12.23). Even though all prefixes are likely to be in the corpus, \wikipedia{}  performance in terms of USPG is low regardless of the decoding strategy. This counter-intuitive behaviour seems to occur mainly due to the tendency of the small model trained on \wikipedia{} to generate endless, unverifiable entity lists, mimicking Wikipedia lists. \ccnews{} leads to the best performance in terms of recall (USPG) for all decoding strategies, but also in terms of precision (USPV) for \tk{} and \dbs.

We did explore several other dimensions, including grammaticality (through a syntactic parser) and relevance (i.e., tf-idf score with the prefix Wikipedia page) during our experiments (see Table \ref{table:otherdimension} in the Appendix). Figure \ref{fig:correlation} reports the Pearson correlation coefficient between supported and verified sentences and these set of metrics. We consider the four runs of the large transformer model reported in Figure \ref{fig:ablmodzize}.
We notice, for instance, that the average log probability of a sentence is positively correlated with verifiability, suggesting that max-likelihood strategies are better suited in this regards. Furthermore, the tf-idf score with the prefix Wikipedia page content is positively correlated with supported sentences. This behaviour is related to the implementation of the fact checker we use, which, by considering exclusively Wikipedia as knowledge source, favours text with a high overlap with the latter. Note, however, that the model was not explicitly exposed to Wikipedia during training (\ie, \ccnews{} does not explicitely include it).

\begin{figure}[t!]
    \centering
    \includegraphics[width=\linewidth]{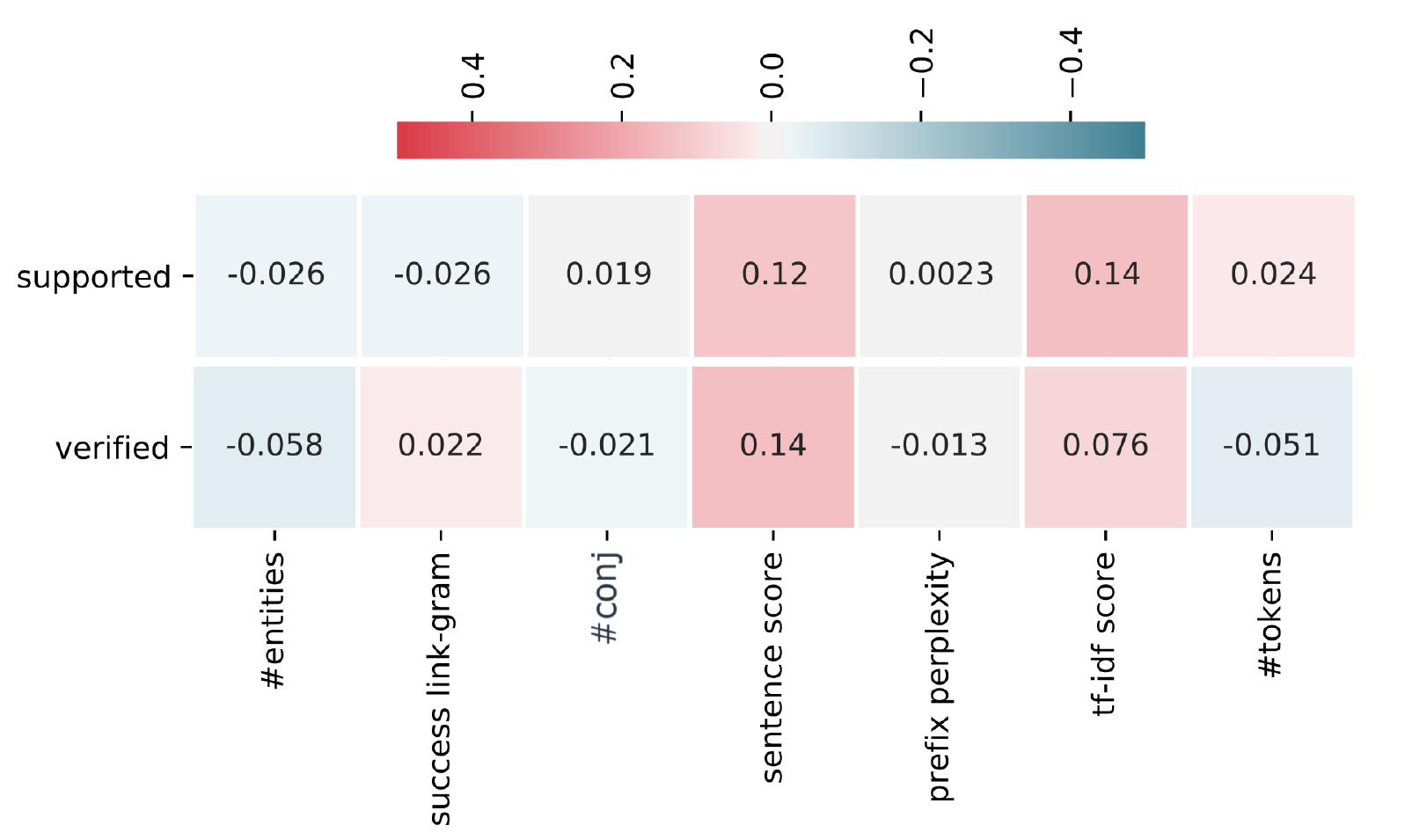}
    \caption{Pearson correlation coefficient for supported/verified sentences (large model) and a set of metrics per sentence: number of entities, if successfully parsed by the Link-Grammar syntactic parser,\footnotemark number of conjunctions in the dependency tree, average token log probability, prefix perplexity, tf-idf score with the prefix Wikipedia page, number of tokens.}
    \label{fig:correlation}
\end{figure}
\footnotetext{\url{abisource.com/projects/link-grammar}}

We report examples of text generated by the large transformer model using different decoding strategies in the Appendix section (Table \ref{table:examplesfinal}).

\paragraph{Human Evaluation} 
We carry out an annotations campaign, where we ask human annotators to fact check generated text. We base the evaluation on a set of 200 prefixes randomly selected from the test set. We consider completions produced by 5 of the generation strategies studied in this paper. We collect 5 annotations per generation. Results, reported in Table \ref{tab:mturk}, confirm our findings: sampling strategies generate text which is less repetitive but also with less supported sentences than in the case of beam search. \dbs{} emerges as a reasonable trade-off between the two, being less repetitive than \bs{} and producing more supported sentences than \tk.
The analysis also highlights how blocking n-grams does not really address the repetitive nature of \bs. Looking at some examples (see Table \ref{table:examplesfinal}) we notice that \bs$_b$ avoids repeating n-grams by introducing superficial, token-level modifications which, most of the time, fail to alternate the underline meaning of the sentence.
In terms of absolute values, precision metrics (\ie, USPV and SPV) are lower than those computed with the automatic fact checker, and recall metrics (\ie, SPG and USPG) higher. This is due to the poor recall performance of the fact checking system - $45.66\%$ for \support{} and $5.78\%$ for \reject. Precision values are $80.89\%$ for \support{} and $52.69\%$ for \reject.
In sum we find that while off-the-shelf, state-of-the-art fact checker systems still leave ample room for improvement, they already serve as a good proxy for ranking pre-trained language models and decoding strategies with respect to the verifiability of the text they generate.

\begin{table}[t!]
\centering

\resizebox{\linewidth}{!}{   

    \begin{tabular}{ c c c c c c c c }

     & \textit{REP} & \textit{NAC} & \textit{UNG} & \textbf{SPG} & \textbf{SPV} & \textbf{USPG} &  \textbf{USPV} \\
    \midrule
    
    \textbf{\tk} & \textbf{16.0} & 3.4 & \textbf{1.4} & 27.2 & 41.36 & 20.1 & 41.16 \\
    \textbf{\greedy}  & 35.6 & \textbf{1.7} & 2.0 & 32.5 & 42.23 & 17.6 & 41.75 \\
    \textbf{\bs} & 38.7 & 8.2 & 1.8 & \textbf{44.6} & \textbf{64.62} & 20.1 & \textbf{65.1} \\
    \textbf{\dbs}  &25.0 & 6.2 & 3.0 & 35.4 & 50.22 & \textbf{23.3} & 50.68 \\
    \textbf{\bs$_b$}  & 31.6 & 9.8 & 4.3 & 38.2 & 56.92 & 19.7 & 58.12 \\

    \bottomrule

    \end{tabular}  }  
\caption{
Results based on human fact checkers, 5 annotations per sentence. Average inter-annotator agreement is 0.66 Cohen's kappa (average majority of $81\%$ for \support, $78\%$ for \reject{} and $65\%$ for \nei). We report the percentage of sentences annotated as repetitions (\textit{REP}), not a claim (\textit{NAC}), ungrammatical (\textit{UNG}), and our verifiability metrics.}
\label{tab:mturk}
\end{table}

\section{Conclusion and Discussion}

We presented a systematic analysis of the verifiability of text generated by a wide range of decoding strategies from large autoregressive language models. We assessed generated sentences with an off-the-shelf automatic fact-checker as well as through human annotations. 
We found that sampling decoding strategies produce text that is less verifiable, but also less repetitive when compared to strategies that consider most likely sequences according to the model distribution. We proposed a hybrid decoding strategy, combining the non-repetitive nature of sampling solutions with the verifiable generation of likelihood-based approaches.

In our analysis, we considered the most viewed Wikipedia pages in 2017 and 2018. Our rationale was that such pages would represent topics that are likely to be highly covered in a random web crawl (e.g., \openwebtext{} and \ccnews). Results (not reported in the paper) with a random set of Wikipedia pages showed lower values in terms of SPG and USPG (\ie, recall metrics). A potential line of future work could be to investigate relationships among training corpora and generation.

We considered each sentence as a single claim to keep our experimental setting clean and avoid noise from an automatic claim extractor. However, some generations contain multiple claims that could be independently assessed. Studying such phenomena is an interesting future direction. 

\bibliography{PINOCCHIO}

\begin{thebibliography}{37}
\expandafter\ifx\csname natexlab\endcsname\relax\def\natexlab#1{#1}\fi

\bibitem[{Bakhtin et~al.(2019)Bakhtin, Gross, Ott, Deng, Ranzato, and
  Szlam}]{bakhtin2019real}
Anton Bakhtin, Sam Gross, Myle Ott, Yuntian Deng, Marc'Aurelio Ranzato, and
  Arthur Szlam. 2019.
\newblock Real or fake? learning to discriminate machine from human generated
  text.
\newblock \emph{arXiv preprint arXiv:1906.03351}.

\bibitem[{Bengio et~al.(2003)Bengio, Ducharme, Vincent, and
  Jauvin}]{DBLP:journals/jmlr/BengioDVJ03}
Yoshua Bengio, R{\'e}jean Ducharme, Pascal Vincent, and Christian Jauvin. 2003.
\newblock A neural probabilistic language model.
\newblock \emph{Journal of machine learning research}, 3(Feb):1137--1155.

\bibitem[{Broscheit(2019)}]{Broscheit2019Investigating}
Samuel Broscheit. 2019.
\newblock Investigating entity knowledge in bert with simple neural end-to-end
  entity linking.
\newblock \emph{Proceedings of the 23rd Conference on Computational Natural
  Language Learning (CoNLL)}.

\bibitem[{Cho et~al.(2019)Cho, Seo, and Hajishirzi}]{cho2019diverse}
Jaemin Cho, Minjoon Seo, and Hannaneh Hajishirzi. 2019.
\newblock Mixture content selection for diverse sequence generation.
\newblock \emph{arXiv preprint arXiv:1909.01953}.

\bibitem[{Dai et~al.(2019)Dai, Yang, Yang, Cohen, Carbonell, Le, and
  Salakhutdinov}]{DBLP:journals/corr/abs-1901-02860}
Zihang Dai, Zhilin Yang, Yiming Yang, William~W Cohen, Jaime Carbonell, Quoc~V
  Le, and Ruslan Salakhutdinov. 2019.
\newblock Transformer-xl: Attentive language models beyond a fixed-length
  context.
\newblock \emph{arXiv preprint arXiv:1901.02860}.

\bibitem[{Devlin et~al.(2018)Devlin, Chang, Lee, and
  Toutanova}]{DBLP:journals/corr/abs-1810-04805}
Jacob Devlin, Ming-Wei Chang, Kenton Lee, and Kristina Toutanova. 2018.
\newblock Bert: Pre-training of deep bidirectional transformers for language
  understanding.
\newblock \emph{arXiv preprint arXiv:1810.04805}.

\bibitem[{Fan et~al.(2018)Fan, Lewis, and Dauphin}]{DBLP:conf/acl/LewisDF18}
Angela Fan, Mike Lewis, and Yann~N. Dauphin. 2018.
\newblock Hierarchical neural story generation.
\newblock In \emph{Proceedings of the 56th Annual Meeting of the Association
  for Computational Linguistics, {ACL}}, pages 889--898.

\bibitem[{Hanselowski et~al.(2018)Hanselowski, Zhang, Li, Sorokin, Schiller,
  Schulz, and Gurevych}]{AtheneFever}
Andreas Hanselowski, Hao Zhang, Zile Li, Daniil Sorokin, Benjamin Schiller,
  Claudia Schulz, and Iryna Gurevych. 2018.
\newblock Ukp-athene: Multi-sentence textual entailment for claim verification.
\newblock \emph{arXiv preprint arXiv:1809.01479}.

\bibitem[{Holtzman et~al.(2019)Holtzman, Buys, Forbes, and
  Choi}]{holtzman2019curious}
Ari Holtzman, Jan Buys, Maxwell Forbes, and Yejin Choi. 2019.
\newblock The curious case of neural text degeneration.
\newblock \emph{arXiv preprint arXiv:1904.09751}.

\bibitem[{Keskar et~al.(2019)Keskar, McCann, Varshney, Xiong, and
  Socher}]{DBLP:journals/corr/abs-1909-05858}
Nitish~Shirish Keskar, Bryan McCann, Lav~R Varshney, Caiming Xiong, and Richard
  Socher. 2019.
\newblock Ctrl: A conditional transformer language model for controllable
  generation.
\newblock \emph{arXiv preprint arXiv:1909.05858}.

\bibitem[{Kiros et~al.(2015)Kiros, Zhu, Salakhutdinov, Zemel, Urtasun,
  Torralba, and Fidler}]{kiros2015skip}
Ryan Kiros, Yukun Zhu, Ruslan~R Salakhutdinov, Richard Zemel, Raquel Urtasun,
  Antonio Torralba, and Sanja Fidler. 2015.
\newblock Skip-thought vectors.
\newblock In \emph{Advances in neural information processing systems 28}, pages
  3294--3302.

\bibitem[{Li et~al.(2016)Li, Monroe, and Jurafsky}]{li2016simple}
Jiwei Li, Will Monroe, and Dan Jurafsky. 2016.
\newblock A simple, fast diverse decoding algorithm for neural generation.
\newblock \emph{arXiv preprint arXiv:1611.08562}.

\bibitem[{Liu et~al.(2019{\natexlab{a}})Liu, Ott, Goyal, Du, Joshi, Chen, Levy,
  Lewis, Zettlemoyer, and Stoyanov}]{liu2019roberta}
Yinhan Liu, Myle Ott, Naman Goyal, Jingfei Du, Mandar Joshi, Danqi Chen, Omer
  Levy, Mike Lewis, Luke Zettlemoyer, and Veselin Stoyanov. 2019{\natexlab{a}}.
\newblock Roberta: A robustly optimized bert pretraining approach.
\newblock \emph{arXiv preprint arXiv:1907.11692}.

\bibitem[{Liu et~al.(2019{\natexlab{b}})Liu, Xu, Winata, and
  Fung}]{liu2019rescoring}
Zihan Liu, Yan Xu, Genta~Indra Winata, and Pascale Fung. 2019{\natexlab{b}}.
\newblock Incorporating word and subword units in unsupervised machine
  translation using language model rescoring.
\newblock \emph{arXiv preprint arXiv:1908.05925}.

\bibitem[{Logan et~al.(2019)Logan, Robert, Nelson, Matthew
  et~al.}]{logan2019barack}
IV~Logan, L~Robert, F~Nelson, E~Matthew, et~al. 2019.
\newblock Barack's wife hillary: Using knowledge-graphs for fact-aware language
  modeling.
\newblock \emph{arXiv preprint arXiv:1906.07241}.

\bibitem[{Melis et~al.(2017)Melis, Dyer, and
  Blunsom}]{DBLP:journals/corr/MelisDB17}
G{\'a}bor Melis, Chris Dyer, and Phil Blunsom. 2017.
\newblock On the state of the art of evaluation in neural language models.
\newblock \emph{arXiv preprint arXiv:1707.05589}.

\bibitem[{Mikolov and Zweig(2012)}]{DBLP:conf/slt/MikolovZ12}
Tomas Mikolov and Geoffrey Zweig. 2012.
\newblock Context dependent recurrent neural network language model.
\newblock In \emph{Proceedings of the fourth IEEE Spoken Language Technology
  Workshop (SLT)}, pages 234--239.

\bibitem[{Nagel(2016)}]{Nagel2016}
Sebastian Nagel.
\newblock
  \url{http://web.archive.org/save/http://commoncrawl.org/2016/10/news-dataset-available/}
  [online]. 2016.
\newblock Accessed: 2019-11-08.

\bibitem[{Niewinski et~al.(2019)Niewinski, Pszona, and
  Janicka}]{niewinski2019tmlab}
Piotr Niewinski, Maria Pszona, and Maria Janicka. 2019.
\newblock Tmlab: Generative enhanced model (gem) for adversarial attacks.
\newblock \emph{arXiv preprint arXiv:1910.00337}.

\bibitem[{Novikova et~al.(2017)Novikova, Du{\v{s}}ek, Curry, and
  Rieser}]{novikova2017we}
Jekaterina Novikova, Ond{\v{r}}ej Du{\v{s}}ek, Amanda~Cercas Curry, and Verena
  Rieser. 2017.
\newblock Why we need new evaluation metrics for nlg.
\newblock \emph{arXiv preprint arXiv:1707.06875}.

\bibitem[{Ott et~al.(2019)Ott, Edunov, Baevski, Fan, Gross, Ng, Grangier, and
  Auli}]{ott2019fairseq}
Myle Ott, Sergey Edunov, Alexei Baevski, Angela Fan, Sam Gross, Nathan Ng,
  David Grangier, and Michael Auli. 2019.
\newblock fairseq: A fast, extensible toolkit for sequence modeling.
\newblock \emph{arXiv preprint arXiv:1904.01038}.

\bibitem[{Paulus et~al.(2017)Paulus, Xiong, and
  Socher}]{paulus2017summarization}
Romain Paulus, Caiming Xiong, and Richard Socher. 2017.
\newblock A deep reinforced model for abstractive summarization.
\newblock \emph{arXiv preprint arXiv:1705.04304}.

\bibitem[{Peters et~al.(2018)Peters, Neumann, Iyyer, Gardner, Clark, Lee, and
  Zettlemoyer}]{DBLP:conf/naacl/PetersNIGCLZ18}
Matthew~E. Peters, Mark Neumann, Mohit Iyyer, Matt Gardner, Christopher Clark,
  Kenton Lee, and Luke Zettlemoyer. 2018.
\newblock Deep contextualized word representations.
\newblock In \emph{Proceedings of the 16th Conference of the North American
  Chapter of the Association for Computational Linguistics: Human Language
  Technologies, NAACL-HLT}, pages 2227--2237.

\bibitem[{Petroni et~al.(2019)Petroni, Rockt{\"a}schel, Lewis, Bakhtin, Wu,
  Miller, and Riedel}]{petroni2019language}
Fabio Petroni, Tim Rockt{\"a}schel, Patrick Lewis, Anton Bakhtin, Yuxiang Wu,
  Alexander~H Miller, and Sebastian Riedel. 2019.
\newblock Language models as knowledge bases?
\newblock \emph{arXiv preprint arXiv:1909.01066}.

\bibitem[{Radford et~al.(2018)Radford, Narasimhan, Salimans, and
  Sutskever}]{radford2018improving}
Alec Radford, Karthik Narasimhan, Tim Salimans, and Ilya Sutskever. 2018.
\newblock Improving language understanding by generative pre-training.

\bibitem[{Radford et~al.(2019{\natexlab{a}})Radford, Wu, Amodei, Clark,
  Brundage, and Sutskever}]{openai2019better}
Alec Radford, Jeff Wu, Dario Amodei, Jack Clark, Miles Brundage, and Ilya
  Sutskever. 2019{\natexlab{a}}.
\newblock Better language models and their implications.
\newblock \url{https://blog.openai.com/better-language-models/}.
\newblock Accessed: 2019-11-08.

\bibitem[{Radford et~al.(2019{\natexlab{b}})Radford, Wu, Child, Luan, Amodei,
  and Sutskever}]{radford2019language}
Alec Radford, Jeffrey Wu, Rewon Child, David Luan, Dario Amodei, and Ilya
  Sutskever. 2019{\natexlab{b}}.
\newblock Language models are unsupervised multitask learners.

\bibitem[{Roberts et~al.(2020)Roberts, Raffel, and Shazeer}]{roberts2020much}
Adam Roberts, Colin Raffel, and Noam Shazeer. 2020.
\newblock How much knowledge can you pack into the parameters of a language
  model?
\newblock \emph{arXiv preprint arXiv:2002.08910}.

\bibitem[{Thorne et~al.(2018)Thorne, Vlachos, Christodoulopoulos, and
  Mittal}]{Thorne_2018}
James Thorne, Andreas Vlachos, Christos Christodoulopoulos, and Arpit Mittal.
  2018.
\newblock Fever: a large-scale dataset for fact extraction and verification.
\newblock \emph{Proceedings of the 16th Conference of the North American
  Chapter of the Association for Computational Linguistics: Human Language
  Technologies (NAACL-HLT)}, pages 809--819.

\bibitem[{Thorne et~al.(2019)Thorne, Vlachos, Cocarascu, Christodoulopoulos,
  and Mittal}]{thorne-etal-2019-fever2}
James Thorne, Andreas Vlachos, Oana Cocarascu, Christos Christodoulopoulos, and
  Arpit Mittal. 2019.
\newblock \href {https://doi.org/10.18653/v1/D19-6601} {The {FEVER}2.0 shared
  task}.
\newblock In \emph{Proceedings of the Second Workshop on Fact Extraction and
  VERification (FEVER)}, pages 1--6, Hong Kong, China. Association for
  Computational Linguistics.

\bibitem[{Vaswani et~al.(2017)Vaswani, Shazeer, Parmar, Uszkoreit, Jones,
  Gomez, Kaiser, and Polosukhin}]{DBLP:conf/nips/VaswaniSPUJGKP17}
Ashish Vaswani, Noam Shazeer, Niki Parmar, Jakob Uszkoreit, Llion Jones,
  Aidan~N Gomez, {\L}ukasz Kaiser, and Illia Polosukhin. 2017.
\newblock Attention is all you need.
\newblock In \emph{Advances in neural information processing systems 30}, pages
  5998--6008.

\bibitem[{Vijayakumar et~al.(2016)Vijayakumar, Cogswell, Selvaraju, Sun, Lee,
  Crandall, and Batra}]{vijayakumar2016diverse}
Ashwin~K Vijayakumar, Michael Cogswell, Ramprasath~R Selvaraju, Qing Sun,
  Stefan Lee, David Crandall, and Dhruv Batra. 2016.
\newblock Diverse beam search: Decoding diverse solutions from neural sequence
  models.
\newblock \emph{arXiv preprint arXiv:1610.02424}.

\bibitem[{Welleck et~al.(2019)Welleck, Kulikov, Roller, Dinan, Cho, and
  Weston}]{DBLP:journals/corr/abs-1908-04319}
Sean Welleck, Ilia Kulikov, Stephen Roller, Emily Dinan, Kyunghyun Cho, and
  Jason Weston. 2019.
\newblock Neural text generation with unlikelihood training.
\newblock \emph{arXiv preprint arXiv:1908.04319}.

\bibitem[{Yang et~al.(2019)Yang, Dai, Yang, Carbonell, Salakhutdinov, and
  Le}]{yang2019xlnet}
Zhilin Yang, Zihang Dai, Yiming Yang, Jaime Carbonell, Ruslan Salakhutdinov,
  and Quoc~V Le. 2019.
\newblock Xlnet: Generalized autoregressive pretraining for language
  understanding.
\newblock \emph{arXiv preprint arXiv:1906.08237}.

\bibitem[{Yu et~al.(2017)Yu, Blunsom, Dyer, Grefenstette, and
  Kocisk{\'{y}}}]{DBLP:conf/iclr/YuBDGK17}
Lei Yu, Phil Blunsom, Chris Dyer, Edward Grefenstette, and Tom{\'{a}}s
  Kocisk{\'{y}}. 2017.
\newblock The neural noisy channel.
\newblock In \emph{Proceedings of the 5th International Conference on Learning
  Representations, ICLR}.

\bibitem[{Zellers et~al.(2019)Zellers, Holtzman, Rashkin, Bisk, Farhadi,
  Roesner, and Choi}]{DBLP:journals/corr/abs-1905-12616}
Rowan Zellers, Ari Holtzman, Hannah Rashkin, Yonatan Bisk, Ali Farhadi,
  Franziska Roesner, and Yejin Choi. 2019.
\newblock Defending against neural fake news.
\newblock \emph{arXiv preprint arXiv:1905.12616}.

\bibitem[{Zhu et~al.(2015)Zhu, Kiros, Zemel, Salakhutdinov, Urtasun, Torralba,
  and Fidler}]{zhu2015aligning}
Yukun Zhu, Ryan Kiros, Rich Zemel, Ruslan Salakhutdinov, Raquel Urtasun,
  Antonio Torralba, and Sanja Fidler. 2015.
\newblock Aligning books and movies: Towards story-like visual explanations by
  watching movies and reading books.
\newblock In \emph{Proceedings of the 15th IEEE international conference on
  computer vision (ICCV)}, pages 19--27.

\end{thebibliography}
\bibliographystyle{acl_natbib}

\newpage
\begin{table}[h!]
\centering
\begin{tabular}{ c c} 
    \toprule
    strategy & best parameters \\
    \midrule
    \tk & k$=2$ \\ 
    \tp & p$=0.4$ \\
    \bs & beam size$=15$\\
    \gdbs & \makecell{ groups$=2$\\ penalty$=0.2$} \\
    \sdbs & penalty$=0.1$ \\
    \dbs & \makecell{ top-k$=100$\\ beam size$=6$;\\ L$=1$} \\
    \bs$_b$ & \makecell{ beam size$=15$\\ blocking order$=20$} \\
    \bottomrule
\end{tabular}  
\caption{Best parameters per decoding strategy.}
\label{tab:best_parameters}
\end{table}
\begin{table*}[htb!]
\centering

\resizebox{0.9\textwidth}{!}{   

    \begin{tabular}{ c cc c c c c cc cc }
    \toprule


      strategy & param  && \makecell{distinct \\ sentences} & \#entities & \#tokens & \#conj && \makecell{  \% success \\ link-gram} &&  \makecell{tf-idf \\ score}  \\
    \midrule
    
\textit{greedy}&     1  &&              2.44 &       2.02 &  18.68 &       1.18 &&      81.30  && 255.04    \\
\midrule

\multirow{3}{*}{\makecell{beam \\ search}} &     6 &&              2.20 &       2.79 &  22.71 &       1.32 && 74.76  && 510.17   \\
 &    12        & &              2.13 &       3.13 &  23.72 &       1.39 &&          71.92  && 565.97   \\
 &    15 &&              2.14 &       3.13 &  23.51 &       1.40 &&     72.41  && 568.96   \\
\midrule

\multirow{3}{*}{\makecell{\tk{}}}&     2 &&              4.63 &       2.43 &  21.92 &       1.26 &&         83.58  && 259.75   \\
 &    10 &&              4.95 &       2.70 &  25.22 &       1.33 &&         78.73  && 246.18   \\
 &   100 &&              4.98 &       2.72 &  27.29 &       1.36 &&          74.91  && 203.19   \\
\midrule

\multirow{4}{*}{\makecell{\tp{}}} & 0.1 &&   2.57 & 2.02 &  18.93 & 1.17 &&  81.10  && 251.27  \\
 &   0.3 && 3.88 &  2.19 &  19.41 & 1.19 && 85.06  && 238.72   \\
 &   0.7 &&              4.90 &       2.59 &  23.76 &       1.27 &&       79.90  && 215.68  \\
 &     1 &&              4.97 &       2.81 &  28.55 &       1.38 &&       70.40  && 162.95  \\
\midrule
 
\multirow{4}{*}{\makecell{delayed \\ beam search}} &     5-6-1 &&              3.74 &       3.01 &  23.39 &       1.22 && 77.23  && 401.76  \\
&    10-6-1 &&              3.95 &       3.04 &  23.95 &       1.23 &&  77.00  && 406.03    \\
&   100-6-1 &&              4.22 &       3.04 &  24.10 &       1.22 &&  76.36  && 399.91   \\
&  1000-6-1 &&              4.30 &       3.03 &  24.62 &       1.21 &&  76.27  && 397.27  \\

    \bottomrule

    \end{tabular}
  }  
\caption{
Analysis of the generated text among different dimensions. We used the prefix in the validation set and the small transformer model trained on \ccnews{} to generate the text. Under the column \textit{params} we report hyper parameters for the decoding strategy considered. For \dbs{} we we have varied only the parameter for \tk{} sampling and we used beam size equal to $6$ and delay length of $1$. 
}
\label{table:otherdimension}

\end{table*}

\section{Appendix}

\subsection{Hyperparameters}
We conduct a parameter sweep on the small transformer model on the validation set. The following table shows the configuration for each decoding strategy that leds to the highest USPG score.

\subsection{Generation Examples}
We reported some examples generated with different strategies in table \ref{table:examplesfinal}.

\definecolor{refuted}{RGB}{238, 46, 47}
\definecolor{supported}{RGB}{0, 140, 72}

\begin{table*}[]
\small
\centering
\resizebox{\textwidth}{!}{   
\begin{tabularx}{\linewidth}{rXX}%
\toprule

\multirow{6}{*}{\rotatebox[origin=c]{90}{prefix}} & \textbf{Princess Margaret, Countess of Snowdon}\textit{. Princess Margaret, Countess of Snowdon,  (Margaret Rose 21 August 1930 - 9 February 2002) was the younger daughter of King George VI and Queen Elizabeth The Queen Mother and the only sibling of Queen Elizabeth II.} & \textbf{Whitey Bulger}\textit{. James Joseph Whitey Bulger Jr. ( September 3, 1929 - October 30, 2018) was an Irish-American organized crime boss and FBI informant who led the Winter Hill Gang in the Winter Hill neighborhood of Somerville, Massachusetts, a city directly northwest of Boston.}\\
\midrule
\multirow{14}{*}{\rotatebox[origin=c]{90}{\textbf{top-k}}}                 & 
\makecell[Xt]{
She is the only surviving member from the Royal Family who was born in the same year as the Queen Mother. \\
\highlight[refuted, draw=black, opacity=0.2]{Princess Margaret, Countess of Snowdon,  (Margaret Rose 21 August 1930 - 9 February 2002) was a British princess who married into the British Royal Family in 1939.} \\
She is a cousin to the Queen, Prince Philip, Prince Charles, Prince William, Prince Harry and Prince Edward.\\
\highlight[supported, draw=black, opacity=0.2]{Princess Margaret, Countess of Snowdon,  (Margaret Rose 21 August 1930 - 9 February 2002) is the younger daughter of King George VI and Queen Elizabeth.} \\
\highlight[refuted, draw=black, opacity=0.2]{She is the only survivor from the Royal Family who was born in the same year as Queen Elizabeth.}} &
\makecell[Xt]{
He was a fugitive for 16 years before his arrest in 2011.\\
\highlight[refuted, draw=black, opacity=0.2]{He was sentenced in 2013 to two life terms plus five years for his role in 11 murders and was released in 2014.}\\
Bulger was found dead in his cell at the U.S. penitentiary in West Virginia.\\
\highlight[refuted, draw=black, opacity=0.2]{He was 89.}\\
\highlight[supported, draw=black, opacity=0.2]{He was serving the life sentences for his role in the infamous Boston crime family.}\\
}
\\
\midrule
\multirow{13}{*}{\rotatebox[origin=c]{90}{\textbf{BS}}}                & \makecell[Xt]{ 
\highlight[supported, draw=black, opacity=0.2]{Princess Margaret, Countess of Snowdon,  (Margaret Rose 21 August 1930 - 9 February 2002) was the younger daughter of King George VI and Queen Elizabeth The Queen Mother and the only sibling of Queen Elizabeth II.}\\ 
\highlight[supported, draw=black, opacity=0.2]{Princess Margaret, Countess of Snowdon,  (Margaret Rose 21 August 1930 - 9 February 2002) was the younger daughter of King George VI and Queen Elizabeth The Queen Mother and the only sibling of Queen Elizabeth II.}\\ 
\highlight[supported, draw=black, opacity=0.2]{Princess Margaret, Countess of Snowdon,  (Margaret Rose 21 August 1930 - 9 February 2002) was the younger daughter of King George VI and Queen Elizabeth The Queen Mother and the only sibling of Queen Elizabeth II.}\\...}
& \makecell[Xt]{ 
\highlight[supported, draw=black, opacity=0.2]{Bulger was one of the FBI's most wanted fugitives for 16 years until he was captured in Santa Monica, California, in 2011.}\\
\highlight[supported, draw=black, opacity=0.2]{He was convicted in 2013 of a litany of crimes, including racketeering, extortion, money-laundering, and murder.}\\
\highlight[supported, draw=black, opacity=0.2]{He was sentenced to two consecutive life sentences plus five years.}\\
He died in federal prison in West Virginia on Tuesday at the age of 89.\\
\highlight[supported, draw=black, opacity=0.2]{Bulger was one of the FBI's most wanted fugitives for 16 years until he was captured in Santa Monica, California, in 2011. }
}\\ 
\midrule
\multirow{11}{*}{\rotatebox[origin=c]{90}{\textbf{DBS}}} & \makecell[Xt]{ 
\highlight[supported, draw=black, opacity=0.2]{\textbf{She} married Antony Armstrong-Jones, a photographer, in 1960.} \\
\textbf{It} was the first marriage for the Queen and the first for Prince Philip, Duke of Edinburgh. \\
\textbf{After} divorcing Armstrong-Jones in 1978, she married Group Captain Peter Townsend in June that same year.\\
\highlight[supported, draw=black, opacity=0.2]{\textbf{She} died at the age of 71 on 9 February 2002.} \\
\textbf{Why} did Princess Margaret marry Antony Armstrong-Jones?
} 
& \makecell[Xt]{ 
\textbf{He} was one of the FBI's most wanted fugitives for 16 years until his capture in 2011. \\
\textbf{Born} in Boston, Whitey Bulger was the son of Irish immigrants. \\
\textbf{After} serving in the U.S. Navy during World War II, Whitey Bulger joined the Irish-American mafia, the Winter Hill Gang, in the early 1950s. \\
\textbf{He} quickly rose through the ranks of the gang, eventually becoming its leader. \\
\textbf{He} was known as "Whitey" because of his light brown hair and blue eyes.
}
\\
\midrule
\multirow{14}{*}{\rotatebox[origin=c]{90}{\textbf{BS$_b$}}} & \makecell[Xt]{ 
\highlight[supported, draw=black, opacity=0.2]{Princess Margaret, Countess of Snowdon (Margaret Rose 21 August 1930 - 9 February 2002) was the eldest daughter of King George VI and Queen Elizabeth The Queen Mother.}\\
Princess Margaret (Margaret Rose 21 August 1930 - 9 February 2002) was the eldest child of King George VI and Queen Elizabeth The Queen Mother.\\
\highlight[supported, draw=black, opacity=0.2]{Princess Margaret, Countess of Snowdon.}\\
 (Margaret Rose 21 August 1930 - 9 February 2002) was the eldest daughter of Queen Elizabeth The Queen Mother.\\
 (Margaret Rose 21 August 1930 - 9 February 2002) was the oldest child of King George VI and Queen Elizabeth The Queen Mother.\\
} 
& \makecell[Xt]{ 
\highlight[supported, draw=black, opacity=0.2]{Bulger was one of the FBI's most wanted fugitives for 16 years \textbf{until} he was captured in Santa Monica, California, in 2011.}\\
\highlight[supported, draw=black, opacity=0.2]{He was convicted in 2013 of a litany of crimes, including racketeering, extortion, money-laundering, and murder.}\\
\highlight[supported, draw=black, opacity=0.2]{He was sentenced to two consecutive life sentences plus five years.}\\
He died in federal prison in West Virginia on Tuesday at the age of 89.\\
Bulger was one of the FBI's most wanted fugitives for 16 years \textbf{before} he was captured in Santa Monica, California, in 2011.
}
\\
\bottomrule
\end{tabularx}
}
\caption{Two examples of text generated with different strategies by the large transformer model. One the left a cherry picked example (in terms of repetitive generation for \bs) while on the right a random one. Sentence refuted by the fact checker are highlighted in red, supported in green.}
\label{table:examplesfinal}
\end{table*}

\subsection{Other metrics}

We explored how decoding strategy affects other dimensions of the generated text. Results are reported in table \ref{table:otherdimension}.
We measure several statistics ovtaer the generated text: 
\begin{itemize}
    \item The average number of distinct sentences for each generated text;
    \item The average number of named entities in each sentence;
    \item The average number of tokens in each sentence;
    \item The average number of conjunctions in the dependency tree of each sentence;
\end{itemize}

To compute the above metrics, we used \emph{spaCy}. In particular we used its tokenizer to split tokens and sentences,  its named entity recognition capability to identify named entities and its dependency parser to count the number of conjunctions.

Furthermore, we analyzed the grammatical correctness of the generated text, counting the success rate of the link-gram parser \footnote{https://github.com/opencog/link-grammar} over the sentences in the generated text.

We also measure the relevance of the generated text against the Wikipedia page that contains the prefix used for the generation.
For this purposes, we compute the \emph{tf-idf} score of the generated text and the related Wikipedia page.

\subsection{Ablation study over delay length}
\begin{figure}[!t]
    \centering
    \includegraphics[scale=0.2]{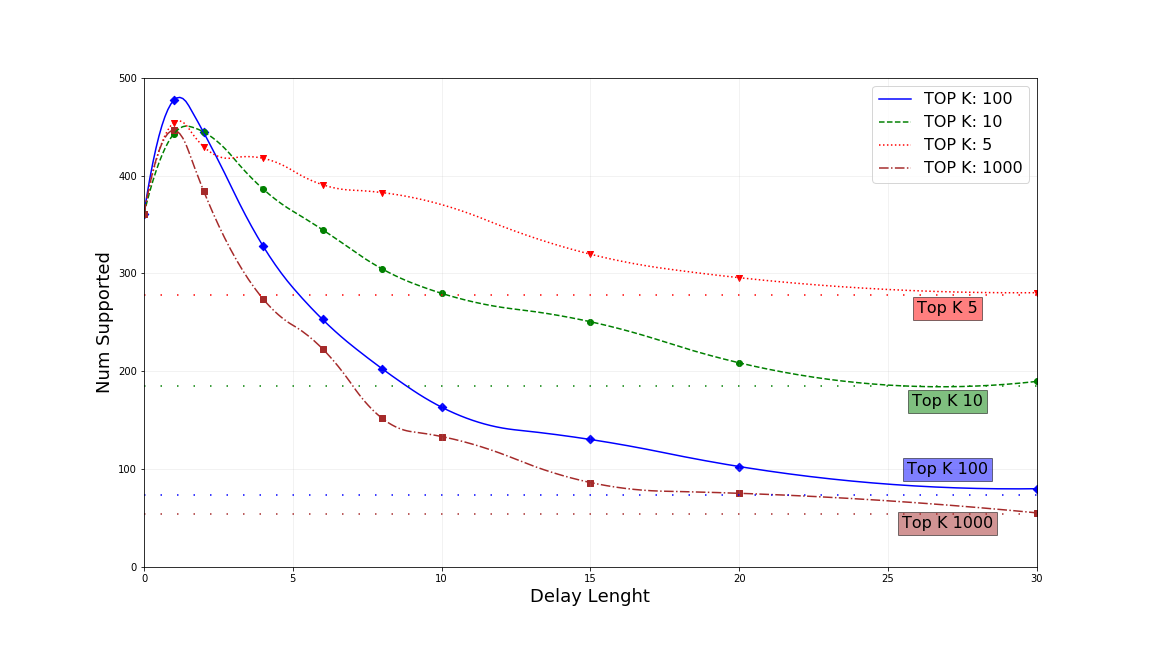}
    \caption{Ablation study over delay length. We report on the x-axis the delay length and on the y-axis the number of distinct supported sentences obtained for each delay length. Horizontal lines represent the value obtained on the validation set using \tk{} decoding strategy. All the generations were performed on the validation set using the small transformer trained on \ccnews{}.}
    \label{fig:ablationdelaylenght}
\end{figure}

We perform an ablation study to measure how the number of supported sentences generated with \dbs{} is affected by the delay length.
We generated text using the prefixes in the validation set using \dbs{} with \tk{} as sampling strategy and with different delay length. Our hypothesis is that using larger delay length the number of supported sentences in the generated text will become close to the one obtained for \tk{}. 
We report the results in figure \ref{fig:ablationdelaylenght}. From the figure it is clear that with larger delay length the number of supported sentences is very close to the one obtained with \tk{}. Moreover, as expected, a short delay length seems to produce a larger number of supported sentences.

\end{document}